\documentclass[letterpaper, preprint, paper,11pt]{AAS}
\usepackage{bm}
\usepackage{amsmath}
\usepackage{amsfonts}
\usepackage{placeins}
\usepackage[colorlinks=true, pdfstartview=FitV, linkcolor=black, citecolor= black, urlcolor= black]{hyperref}
\usepackage{overcite}
\usepackage{footnpag}			      	
\usepackage{comment}
\usepackage{float} 
\usepackage{siunitx} 
\usepackage{subcaption}
\usepackage{soul}
\usepackage{booktabs}

\PaperNumber{25-778}

\begin{document}
\title{REAL-TIME TESTING OF SATELLITE ATTITUDE CONTROL WITH A  REACTION WHEEL HARDWARE-IN-THE-LOOP PLATFORM}

\author{
Morokot Sakal\thanks{Ph.D. Student, Aerospace, Physics and Space Sciences Department, Florida Institute of Technology},  
George Nehma\footnotemark[1],  
Camilo Riano-Rios\thanks{Assistant Professor, Aerospace, Physics and Space Sciences Department, Florida Institute of Technology},  
and Madhur Tiwari\footnotemark[2]
}

\date{Florida Institute of Technology, 150 W. University Blvd., Melbourne, FL, 32901}

\maketitle

\section{abstract}
We propose the Hardware-in-the-Loop (HIL) test of an adaptive satellite attitude control system with reaction wheel health estimation capabilities. Previous simulations and Software-in-the-Loop testing have prompted further experiments to explore the validity of the controller with real momentum exchange devices in the loop. This work is a step toward a comprehensive testing framework for validation of spacecraft attitude control algorithms. The proposed HIL testbed includes brushless DC motors and drivers that communicate using a CAN bus, an embedded computer that executes control and adaptation laws, and a satellite simulator that produces simulated sensor data, estimated attitude states, and responds to actions of the external actuators. We propose methods to artificially induce failures on the reaction wheels, and present related issues and lessons learned. 

\section{Introduction}
Reaction Wheel (RW) arrays are a crucial means for attitude control on many satellites due to their ability to precisely execute the control actions required for attitude maneuvers via exchange of angular momentum\cite{Markley2010}. As such, it is a highly critical subsystem that incorporates levels of redundancy in the case of actuator faults. Extensive research has been made to enhance the capabilities of attitude controllers so that they minimize the hardware cost and computational load as much as possible whilst being able to guarantee tracking \cite{Rev-hassrizal2016survey, Rev-OVCHINNIKOV2019100546, Rev-HASAN2022100806, Rev-AHMEDKHAN20223741}. Methods such as Sliding Mode Control (SMC) \cite{Li2022-nt}, Model Predictive Control (MPC) \cite{MPC-MIRSHAMS2016140, MPC-IANNELLI2022401}, neural networks \cite{rs13122396} as well as a variety of adaptive controllers \cite{RIANORIOS2020189, Sun2021, Xie2023-qc, M_Sadigh2023-pl, Wang2020-mi} have been proposed for attitude control, each with varying capabilities and degrees of success. The advantage of adaptive controllers is their ability to compensate for uncertainties in the system dynamics, whilst maintaining stability. 

In our previous work \cite{nehma2025adaptive}, we proposed a method to simultaneously learn the health of a satellite's RWs and attitude tracking under scenarios of failing or degraded RW(s). Our method involves a Lyapunov-based adaptive controller with an integral concurrent learning (ICL)-based adaptive update law that ensures convergence of the estimated health of the RWs once a finite excitation condition is met. This controller was shown to guarantee exponential convergence of error states and RW health estimates. We demonstrated via MATLAB/Simulink-based numerical simulations that the controller correctly estimated the health of each RW and performed the alternating attitude tracking reference required by the mission. We presented simulations with arrays of 4 and 6 RWs, a varying number of degraded RWs, and varying levels of degradation. A Software-in-the-Loop (SIL) test of the controller was conducted 
to test the real-time capability with embedded hardware (NVIDIA Jetson Nano), achieving comparable results. 

The promising SIL experiments have prompted us to continue with a Hardware-in-the-Loop (HIL) test to validate the adaptive controller in more realistic conditions, including hardware limitations, sensor noise, and communication latency that are difficult to replicate in simulation. 
In addition, we aim to use this controller to demonstrate and validate our envisioned modular testbed, currently under development, at the Space Vehicle and Robotics (SVR) lab at the Florida Institute of Technology. 

HIL testing is a vital aspect in the development of real, flight-ready controllers and actuator-driven systems, as it is a mission-critical system that can afford little to no failures. The sim-to-real gap is large and poses a number of issues when developing control systems, the greatest of which is the ability to model the behavior of the actuating system in simulation as close to the real behavior as possible. 
RW dynamics, although can be approximated in simulation, are often hard to be precisely emulated due to their physical complexity and interaction with motor drivers, sensors, inner control loops, among others. Hence, the need to verify the health estimation capability of a controller using real hardware is of great importance. 


The contributions of this paper are two-fold:
\begin{itemize}
    \item We develop a Hardware in the Loop testing architecture to verify the actuator health estimation performance of the adaptive controller with real RWs.
    \item We highlight the problems encountered during HIL testing of a RW array, and provide solutions and discussion on each.
\end{itemize}

This paper is organized as follows. First, we provide an overview of the adaptive controller that is being tested in this HIL setup. Then we present the architecture design of the testbed. 
The next sections describe the experimental setup and procedures to integrate each component to realize the test. Finally, we discussed the results and concluded with future work. 

\section{Adaptive ICL Controller}

Since the focus of this paper is to highlight and analyze the HIL testing for this adaptive controller, this section only briefly overviews the design and architecture of our controller. For further, more detailed explanation and derivation of the controller and its stability analysis, we refer the reader to our previous paper \cite{nehma2025adaptive}. 

The Equations of Motion for the attitude of a spacecraft with ~$N$ RWs are given as 
\begin{align}
    \label{eq:Eulers_law}
	J\boldsymbol{\dot{\omega}} &= -\boldsymbol{\omega} \times \left(J\boldsymbol{\omega} + J_{RW}G\boldsymbol{\Omega} \right) + G\Phi\boldsymbol{u}  \\ 
        \boldsymbol{\dot{\sigma}} &= \frac{1}{4}\left[\left( 
1-\boldsymbol{\sigma^T\sigma}\right)I_3 + 2\sigma^{\times}+2\boldsymbol{\sigma\sigma^T}\right]\boldsymbol{\omega},\label{eq:MRP_kinem}
\end{align}

where $\boldsymbol{\omega}\in\mathbb{R}^3$ is the spacecraft angular velocity expressed in the body coordinate system, $\boldsymbol{\sigma\in\mathbb{R}^3}$ is the vector of Modified Rodrigues Parameters (MRP) that represent the orientation of the spacecraft with respect to the inertial frame, $\boldsymbol{\Omega}=[\Omega_1, \Omega_2, \cdots, \Omega_N]^T~\in\mathbb{R}^N$ is a vector containing the $N$ RW angular velocities, $\boldsymbol{u}=-J_{RW}\boldsymbol{\dot{\Omega}}=[u_1, u_2, \cdots, u_N]^T ~ \in \mathbb{R}^N$ is the control input that represents the torque applied by each RW, $J\in\mathbb{R}^{3\times 3}$ is the total inertia matrix, $J_{RW}\in\mathbb{R}_{>0}$ is the inertia of the flywheels about their spin axis, $\Phi=diag\{\phi_1,~\phi_2,~\cdots,~\phi_N\}\in\mathbb{R}^{N
\times N}$ is the uncertain RW health matrix, $G=\{\boldsymbol{\hat{s}_1, \hat{s}_2, \cdots, \hat{s}_N}\}\in\mathbb{R}^{3\times N}$ is the RWA configuration matrix, and $\boldsymbol{\hat{s}_i}\in\mathbb{R}^3$ is the direction of the $i^{th}$ RW's spin axis expressed in the body coordinate system. The matrix $I_m\in\mathbb{R}^{m\times m}$ represents an identity matrix of dimension $m\times m$, and $a^{\times}\in\mathbb{R}^{3\times 3}$ is a the skew-symmetric matrix built with the vector $\boldsymbol{a}=[a_1, a_2, a_3]^T\in\mathbb{R}^3$.

The designed auxiliary control law $\boldsymbol{u_d}$ that stabilize the spacecraft attitude is

\begin{equation}
    \label{eq:ctrl_law}
    \boldsymbol{u_d}=\boldsymbol{\omega}\times \left(J\boldsymbol{\omega}+J_{RW}G\boldsymbol{\Omega}\right) + J\Tilde{R}\boldsymbol{\dot{\omega}_d}-J\Tilde{\omega}^{\times}\Tilde{R}\boldsymbol{\omega_d}+4JB^{-1}\left[-\frac{1}{4}\dot{B}\boldsymbol{\tilde{\omega}}-\alpha\boldsymbol{\dot{\sigma}_e}-K\boldsymbol{r}-\beta\boldsymbol{\sigma_e}\right],
\end{equation}
where $\beta\in\mathbb{R}_{>0}$ is a constant control gain, $\alpha\in\mathbb{R}^{3\times 3}$ is a symmetric, positive definite control gain matrix, $\Tilde{R}\in\mathbb{R}^{3\times 3}$ is a rotation matrix between the spacecraft desired and body frames, $\boldsymbol{\sigma_e\in\mathbb{R}^3}$ is the error MRP and $\boldsymbol{r}=\boldsymbol{\dot{\sigma}_e}+\alpha\boldsymbol{\sigma_e}\in\mathbb{R}^3$ is a modified error state.

The torque commands to be sent to the RWs, $\boldsymbol{u}$, are recovered as
\begin{equation}
    \label{eq:u_real} \boldsymbol{u}=\left(G\hat{\Phi}\right)^{\dag}\boldsymbol{u_d}, 
\end{equation}
where $(\cdot)^{\dag}$ is the Moore-Penrose pseudo-inverse of $(\cdot)$ and $\hat{\Phi}$ can be obtained by numerically integrating the adaptation law given by:

\begin{equation}
    \label{eq:adapt_law}
    \boldsymbol{\dot{\hat{\theta}}}=\mathrm{proj}\left\{\frac{1}{4}\Gamma Y^T\left(J^{-1}\right)^TB^T\boldsymbol{r}+\Gamma K_1\sum_{i=1}^{N_s}\mathcal{Y}_i^T\left(J\boldsymbol{\omega}(t)-J\boldsymbol{\omega}(t-\Delta t)+\boldsymbol{\mathcal{U}_i}-\mathcal{Y}_i\boldsymbol{\hat{\theta}}\right)\right\}, 
\end{equation}
where ~$\hat{\boldsymbol{\theta}}\in\mathbb{R}^N$ is the estimate of the vector containing RWs' health factors ~$\boldsymbol{\theta}=[\phi_1, \phi_2, \cdots, \phi_N]^T$, $\Gamma,~K_1,~\in\mathbb{R}^{N\times N}$ are constant, positive definite adaptation gain matrices, $\boldsymbol{\mathcal{U}_i}\in\mathbb{R}^3$ and $\mathcal{Y}_i\in \mathbb{R}^{3\times N}$ are integrals of input-output data terms, and the matrix $B\in\mathbb{R}^{3\times 3}$ is a matrix used in the MRP kinematics\cite{Schaub2018}.

\section{Testbed Architecture Design}
\begin{figure}[h]
	\centering
    \includegraphics[width=0.9\linewidth]{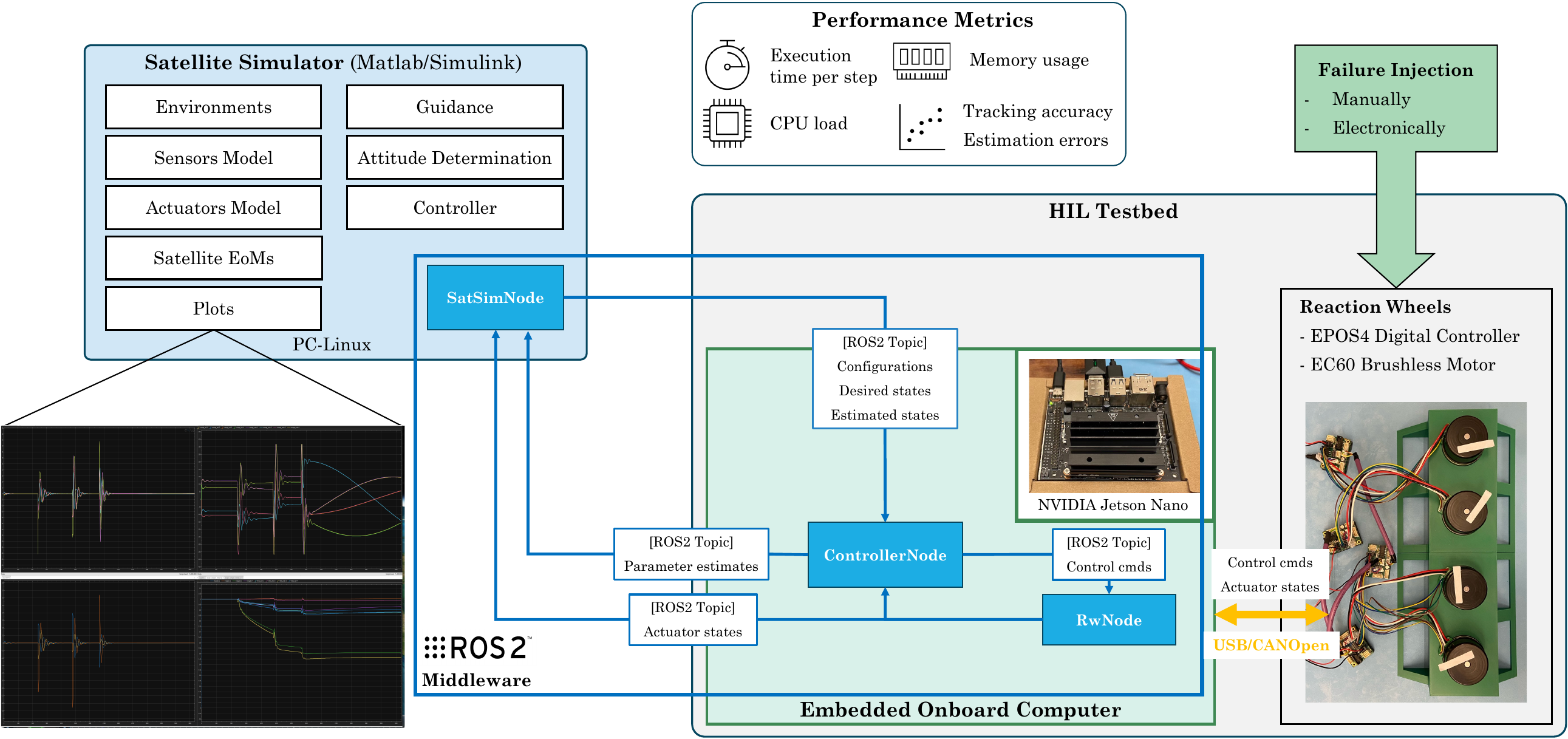}
	\caption{Overview of Testbed Architecture. }
	\label{fig:testbed_arch}
\end{figure}

Figure \ref{fig:testbed_arch} illustrates the overall architecture of the HIL testbed developed for the experiment. 
This architecture extends the already tested and proven SIL setup\cite{nehma2025adaptive} with ROS2 middleware used as for communication between computers in a local network\cite{doi:10.1126/scirobotics.abm6074}. 
The HIL testbed includes a Satellite Simulator implemented in Matlab/Simulink running on a Linux computer, an NVIDIA Jetson Nano embedded computer hosting various software and algorithms undertest, and the RW testbed, which consists of hardware components including four Maxon EC 60 flat brushless motors\cite{maxon_EC60flat} emulating RWs, each controlled by a digital position controller EPOS4 Compact 50/5CAN\cite{maxon_EPOS4}communicating via USB/CAN bus, and an artificial fault injection mechanism used to simulate various RW fault scenarios. 

As shown in Figure \ref{fig:testbed_arch}, the ROS2 nodes in the HIL test setup utilize the publisher/subscriber model. 
Satellite Simulator Node (SatSimNode) runs the simulation of the satellite dynamics, environment, actuator models, and sensor models. 
The sensor models, including sun sensors, magnetometers, and gyroscopes, are modeled by adding noise and realistic sampling rates. 
The simulated rates are: gyroscope 10 Hz, magnetometer 2 Hz, and sun sensors 2 Hz. 
The noisy sensor outputs are processed by the attitude determination block, which employs an Extended Kalman Filter (EKF) based on multiplicative quaternion\cite{crassidis2011optimal} to estimate the attitude states, i.e., satellite inertial angular velocity $\boldsymbol{\omega}$, and quaternion $\rightarrow$ MRPs $\boldsymbol{\sigma}$. 
The SatSimNode provides desired attitude states, satellite configuration parameters, and estimated states data. It also receives the controller outputs and actuator states data, i.e., RW angular velocity and current measurements back from the testbed. 

The embedded computer operates two ROS2 nodes: the ControllerNode and the RwNode, each designated for a specific function. 
The ControllerNode is implemented in Python (rclpy) while the RwNode is implemented in C++ (rclcpp).  
Using the estimated attitude states, desired spacecraft states, and satellite configuration, the ControllerNode calculates the required torque command. 
Depending on the test configuration, the torque command is converted to angular velocity or current commands. The ControllerNode publishes this command to the network via a ROS2 topic. The RwNode then subscribes to this command topic and forwards the commands (using the EPOS Command Library written in C++) to the digital controller via a USB/CANopen interface. 
Simultaneously, the RwNode acts as a publisher and delivers angular velocity and current measurement as the actuator states at the rate of 20 Hz back to the network, which is received by both the ControllerNode and SatSimNode. The final step is to create the closed-loop HIL system and ensure synchronization between the simulation model and the physical hardware. 

The testbed is built with a modular design. It is possible to swap in new actuators, sensors, or embedded computers with minimal changes to the overall architecture. 
It is flexible to make the transition between testing modes, from pure simulation, also called Model-in-the-Loop (MIL), to SIL, or HIL modes. 
This architecture is scalable to support future upgrades, such as air-bearing tests, satellite integration and testing campaigns, and comprehensive end-to-end system tests. 


\section{Experimental Setup}
\begin{figure}[h]
	\centering
    \includegraphics[width=0.6\linewidth]{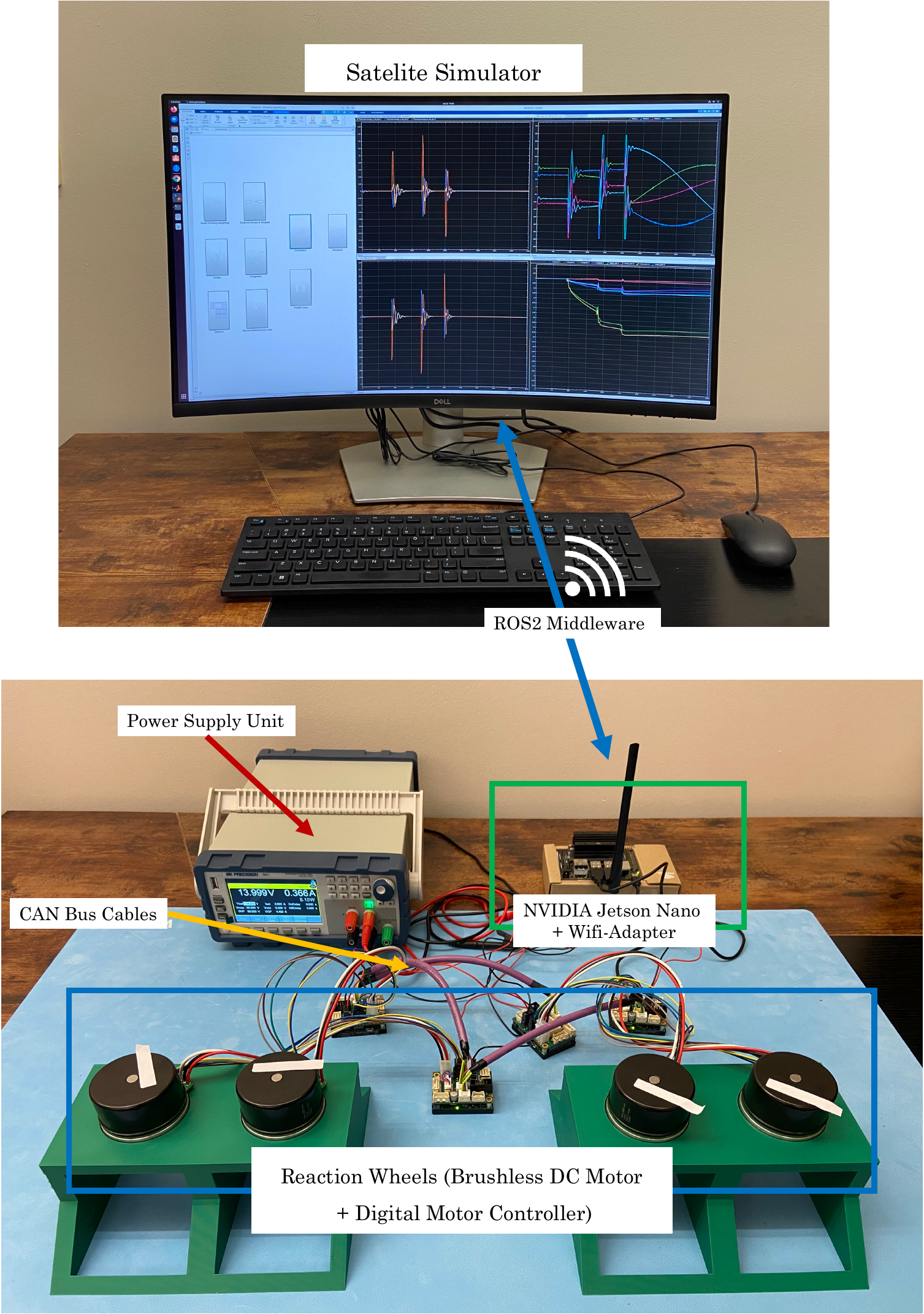}
	\caption{HIL Experimental Setup}
	\label{fig:ros2_rw_setup}
\end{figure}

Figure \ref{fig:ros2_rw_setup} shows the experimental setup used to validate the controller and its RW health estimation capability with actual hardware. 
The NVIDIA Jetson Nano is connected to one digital motor controller via USB interface, and the remaining motor drivers are interconnected through a CAN bus, using the CANopen protocol. 
A dedicated power supply unit is installed to power the electronics. 
The Jetson Nano communicates with MATLAB/Simulink wirelessly in a local network and exchanges data using the ROS2 middleware. 

Performing a HIL test introduces a number of complexities that are either ignored or not present in simulation, one of which is the assumption that all states of the RW are available to the controller, or whether the inner controller loop of the motor controller is closed via current or angular velocity. These considerations greatly impact the design of the controller and whether a real-time HIL test is feasible or representative. 

As a first proof of concept, we artificially induce failure through manipulation of the control command being sent to each RW, as is common practice in many recent literature \cite{doi:10.2514/6.2011-1472, 10752510, doi:10.2514/6.2012-5040, app8101893, 8868202, 7739739, s19214721, 7139933, article, article2}. Testing full failure of the RW is simple, the power stage of the failing RW can be disconnected, whilst other failures can be induced by manipulating the amount of effort that is sent as command to the motor controller with respect to the effort output by the main attitude controller. This is normally a scaling factor that is applied to the required control effort. 


\section{Experimental Procedure}

\begin{figure}[h]
	\centering
    \includegraphics[width=0.8\linewidth]{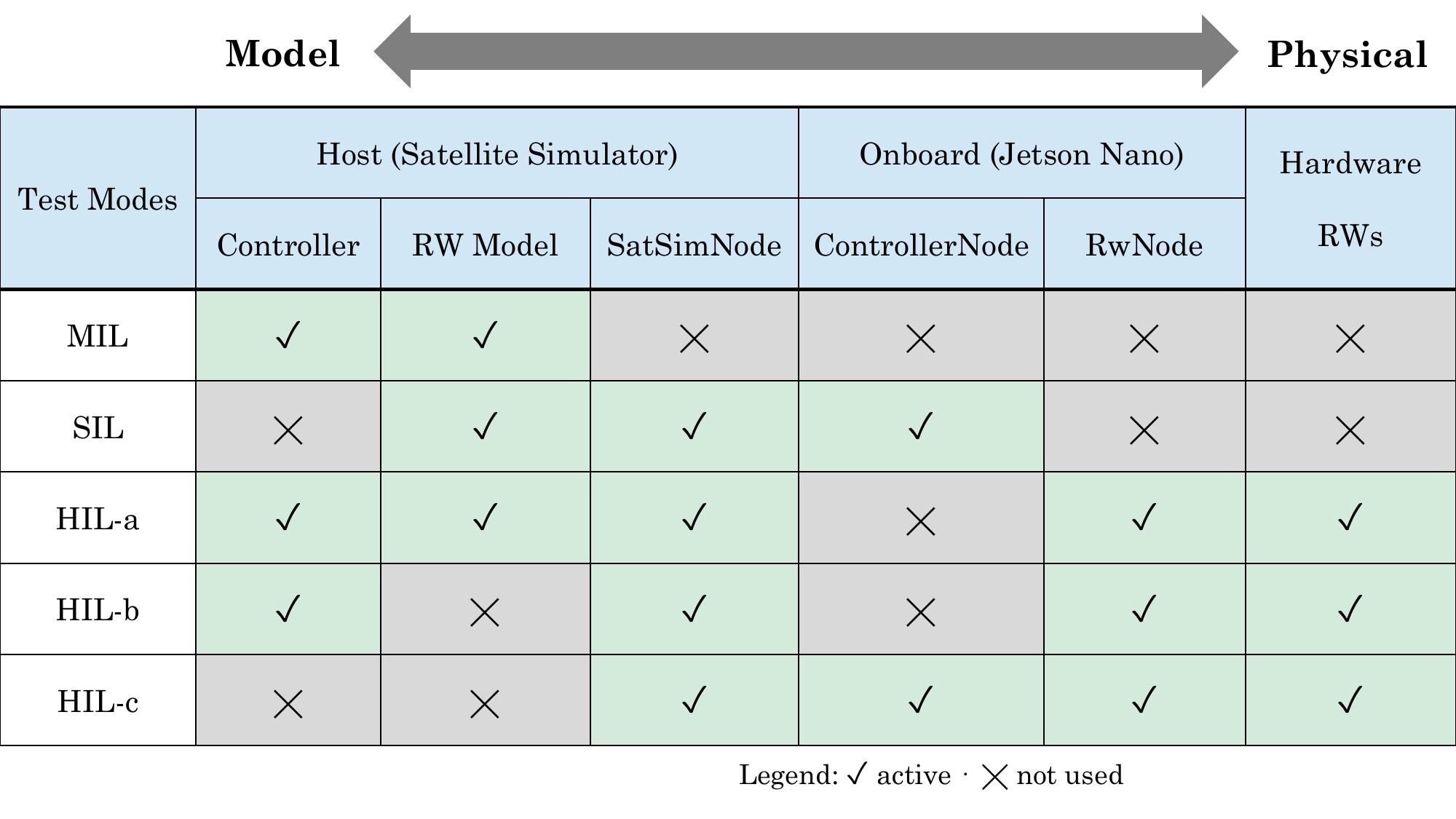}
	\caption{Experimental Procedures}
	\label{fig:exp_step}
\end{figure}

The process of transferring a completely simulated system to one where the actuators and sensors are real hardware connected to the integrated, embedded controller and state estimator is a large jump to complete altogether. As such, we break down the process of implementing the physical hardware in the loop in multiple steps to ensure accuracy and reliability. The first step, as seen in our previous work \cite{nehma2025adaptive} included the SIL testing of the controller executed on an embedded computer. In order to completely verify the HIL tests, the breakdown of how we integrated certain components is shown in Figure \ref{fig:exp_step}. After the SIL test, which verifies that the embedded computer can run the control algorithm accurately and efficiently as compared to the simulation, we begin by adding the reaction wheels, with their CAN bus motor drivers and integrated sensors in the loop. 
The HIL test was further broken down to three subsequent steps: HIL-a, HIL-b, and HIL-c. The goal of each test is to compare the behavior of the model in the Satellite Simulator to a hardware implementation. Once verified, the model in the Satellite Simulator can be gradually removed and replaced with a hardware implementation. 

\subsection{HIL-a: Testing Reaction Wheel Response}

\paragraph{Current Commands}
In order to create the most realistic simulated model of the RW and motor driver set up, control torque commands calculated by the adaptive controller are passed through a transfer function that is designed to mimic the behavior of the motor driver and RW operating in closed-loop torque control. As such, the first test is to pass the control commands, converted to current via the torque constant for the brushless motor \cite{maxon_EC60flat}, from the simulator to the RW and motor drivers and then observe their behavior in comparison to the simulated RWs. Hence, the simulation propagates the equations of motion (EOM) and RW dynamics, with the torque commands being fed in parallel through the ROS2 network to the real RW and motor drivers, but maintaining the control-loop closed with the simulated RWs. The built-in current sensor reports back the actual current, and the Hall-effect sensor measures the angular velocity of each RW. 

As the simulation runs, we compare the output torques and angular velocities between the simulated and real RWs to assess the differences between them. Parameters of the RW such as inertia, mass and torque constant were taken from the datasheet of the brushless motors and included in the simulated RWs. 
Table \ref{tab:sat_params} shows satellite configuration parameters that were adjusted to reflect the physical hardware specifications used in the HIL test. 

\begin{table}[t]
\centering
\caption{Satellite Configuration Parameters (Simulation vs. HIL )}
\begin{tabular}{l|c|c|c}
\toprule
\textbf{Parameter} & \textbf{Simulation} & \textbf{HIL} & \textbf{Units} \\
\midrule
$m$ & 65 & 20 & kg \\
$J$ & $\mathrm{diag}\{0.44,\,0.70,\,0.70\}$ & $\mathrm{diag}\{0.30,\,0.42,\,0.42\}$ & kg$\cdot$m$^{2}$ \\
$G$ & \multicolumn{2}{c|}{$\begin{bmatrix}
0.5774 & -0.5774 & 0.5774 & -0.5774 \\
0.5774 &  0.5774 & -0.5774 & -0.5774 \\
0.5774 &  0.5774 &  0.5774 &  0.5774
\end{bmatrix}$} & -- \\
Max RW torque & $20 \times 10^{-3}$ & $50 \times 10^{-3}$ & N$\cdot$m \\
Max $\Omega$ & $1.04 \times 10^{3}$ & $3.66 \times 10^{2}$ & rad/s \\
\bottomrule
\end{tabular}
\label{tab:sat_params}
\end{table}

The RW motor drivers have two modes of operation: torque (current) command and angular velocity command. In order to reduce complexity and added calculations that may introduce unwanted behaviors, we decided to control the RW's based on a given torque command. In order to do this on the RW hardware, a current command would be delivered to the motor drivers which in turn would use their internal PID feedback controller to track this command. We derive the current command from the torque required by the controller

\begin{equation}
    \label{eq:current_cmd}
    I_{cmd}=K_t\cdot\tau_{cmd}
\end{equation}

where, $K_t$ is the torque constant of the RW motor, given in the datasheet.

\begin{figure}[h]
    \begin{subfigure}[t]{0.45\textwidth}
        \centering
        \includegraphics[width=\linewidth]{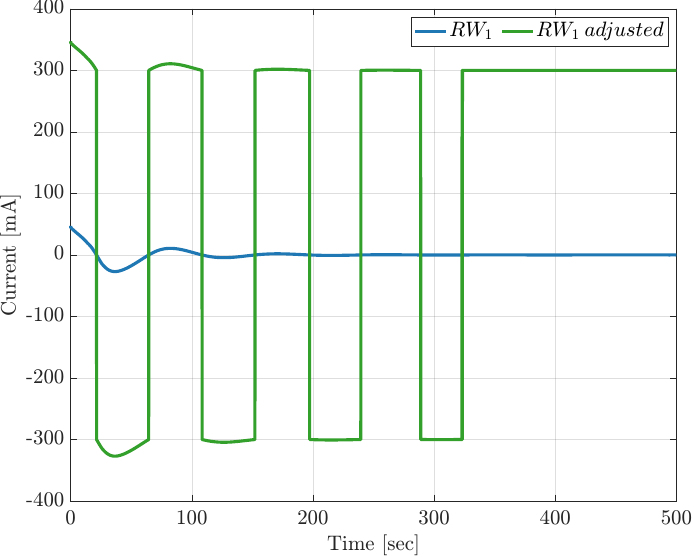}
        \caption{Controller current commands and deadband adjusted current commands sent to RWs.}
        \label{fig:HIL_deadband_current}
    \end{subfigure}
    \hfill
    \begin{subfigure}[t]{0.45\textwidth}
        \centering
        \includegraphics[width=\linewidth]{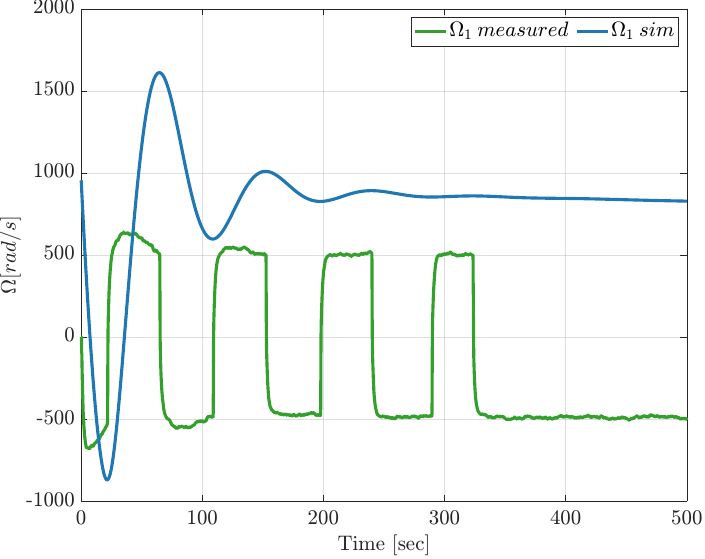}
        \caption{Angular Velocity of Simulated RW and real RW with deadband}
        \label{fig:HIL_deadband_omega}
    \end{subfigure}

    \caption{Current commands and angular velocity plots showing the deadband issue when generating torque commands from the adaptive controller.}
    \label{fig:HIL_deadbands}
\end{figure}

As seen in Figure \ref{fig:HIL_deadband_current}, the primary issue with this method of control commands is that the RWs have a current deadband of around 300mA, throughout which any current commanded would result in no output from the RWs. This is not an uncommon issue \cite{wertz2012spacecraft, carrara2011speed, carrara2013torque} but in our case, because the desired torque commands were generally small and would eventually converge to zero, a majority fell within this deadband, as such the RWs were unable to actuate. A common solution to this issue is to kickstart the RWs when the simulation starts to break through the deadband, and although this momentarily helped alleviate the deadband issue, the motors quickly became unresponsive. Another solution that proved unsuccessful was to append the command signal with the sign of the command multiplied by the deadband region, $I_{cmd} = I_{cmd}+\mathrm{sign}(I_{cmd})\cdot300$. As evident in Figures \ref{fig:HIL_deadband_omega} this resulted in large jumping in the RW angular velocity and did not track the simulated angular velocity closely. As such, our next solution to this problem was to convert the output torque commands of the controller into angular velocity commands.

\paragraph{Velocity Commands}

In order to convert the torque commands that are output from the adaptive controller into velocity commands that are acceptable to the RW motor drivers, the following steps must occur. First, the commanded simulated torque for each RW is saturated by the maximum torque that is possible by each RW, given in its respective datasheet. Then, the torque command is divided by the inertia of the wheel to obtain angular acceleration

\begin{equation}
    \label{eq:vel_cmd}
    \dot{\Omega}_{cmd}=\frac{\tau_{cmd}}{J_{RW}},
\end{equation}
and the velocity command obtained using the forward-Euler integration algorithm which proved to be sufficient. Velocity saturation was also applied to ensure commands within the motor specs. These velocity commands can be tracked by the RWs and do not suffer from the deadband issue. 

\begin{figure}[h]
    \centering
    \includegraphics[width=0.8\linewidth]{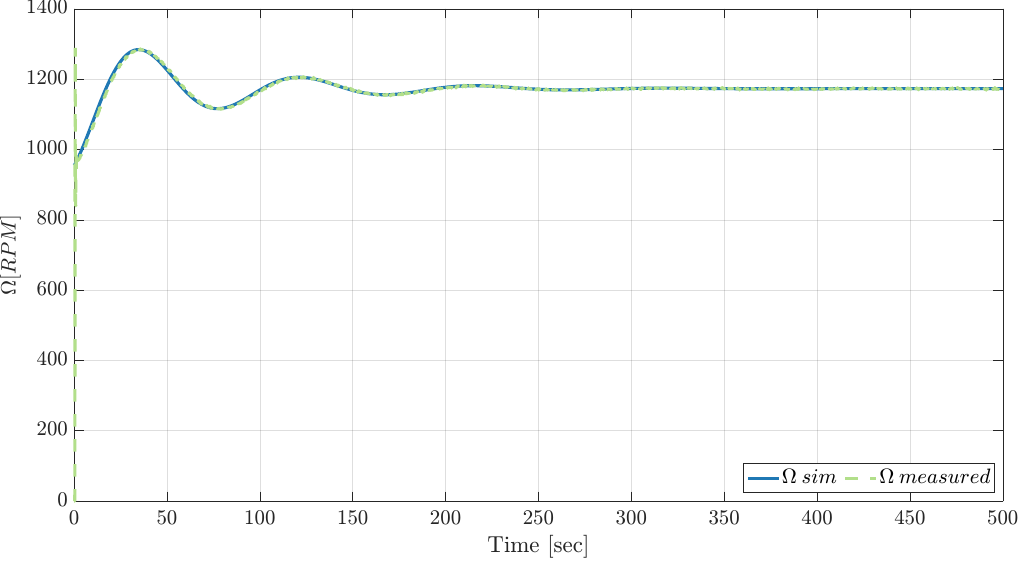}
    \caption{Comparison of simulated and physical RWs' angular velocities under angular velocity command.}
    \label{fig:v_cmd_ang_vel}
\end{figure}

To verify the correct operation of the RWs with the velocity commands generated from the simulator, a similar experiment was performed where the main control loop is still closed with the simulated RWs that receive torque commands, and the corresponding computed velocity commands were also sent to the physical RWs. Figure \ref{fig:v_cmd_ang_vel} demonstrates that although the measured velocity signal includes a small amount of noise, its profile closely follows that of the simulated RW.

\subsection{HIL-b: Adding Physical RWs in the Main Control Loop}

Once it was verified that the simulated and real RWs were behaving similarly for given control commands, the next step in achieving a complete HIL test was to feed the angular velocities and angular accelerations from the real RWs into the simulation, as required to propagate the EoMs as in Equation \eqref{eq:Eulers_law}. This was a major step in the process as it introduced a number of issues that needed to be addressed to bridge the sim-to-real gap. 


Dealing with noisy measurements from the RWs' angular velocities and currents, especially velocity measurements near zero speed, a known issue related to Hall-effect sensors \cite{shigeto2018development}, added to the need to compute the corresponding angular accelerations to be able to propagate the EoMs, became an important issue.  We attempted two approaches to compute the angular acceleration (a) by directly converting the RWs' current measurements to angular acceleration and (b) through numerical differentiation of the velocity measurements, then using low-pass filters to smooth out the signals. 
For the approach (a), we found a similar issue earlier with the deadband of the current measurement, which resulted in an incorrect mapping between current and angular acceleration. 
For approach (b), it introduced a phase shift (delay) in the signal, which is quite significant to be fed into the EoMs, resulting in an unstable system.

\begin{figure}[h]
    \begin{subfigure}[t]{0.45\textwidth}
        \centering
        \includegraphics[width=\linewidth]{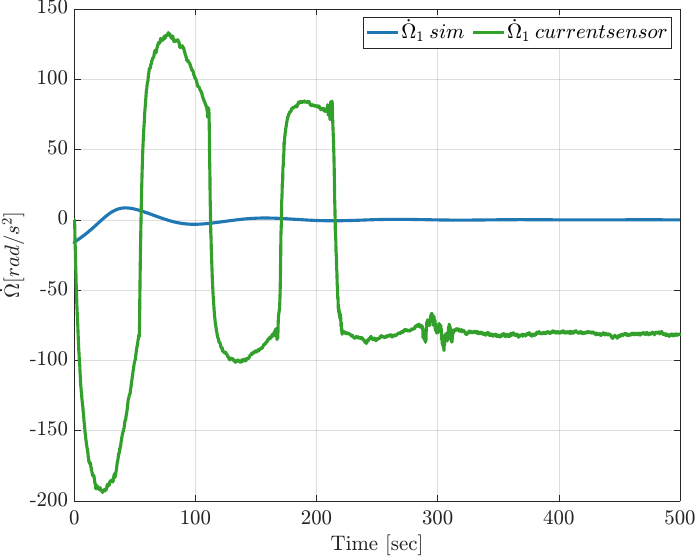}
        \caption{}
        \label{fig:accel_current}
    \end{subfigure}
    \hfill
    \begin{subfigure}[t]{0.45\textwidth}
        \centering
        \includegraphics[width=\linewidth]{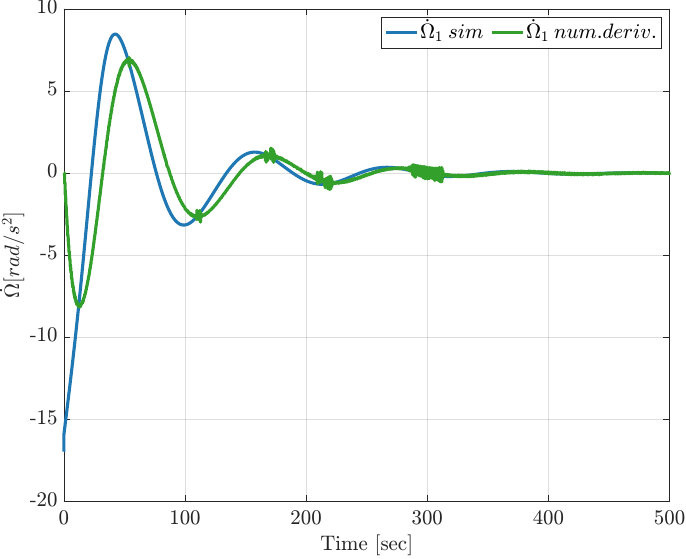}
        \caption{}
        \label{fig:accel_nd}
    \end{subfigure}

    \caption{Wheel acceleration from current sensors and numerical derivative}
    \label{fig:wheel_accel}
\end{figure}


\begin{figure}[h]
    \centering
    \includegraphics[width=0.6\linewidth]{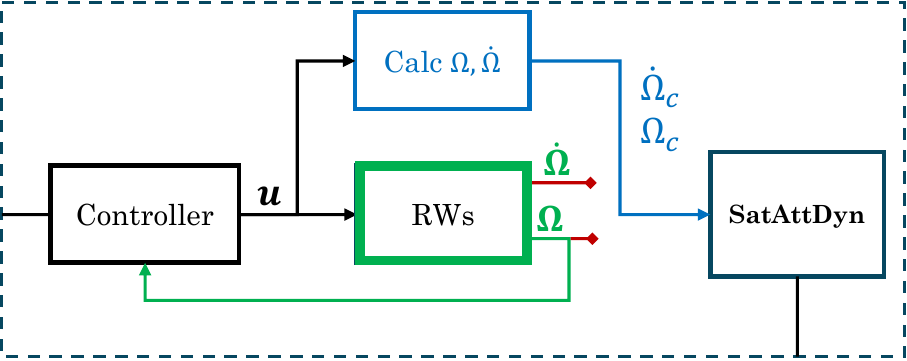}
    \caption{Implementation of the calculation blocks to feed RW states to the EoMs}
    \label{fig:hilb_post}
\end{figure}

To mitigate the issue with noisy sensor data, we fed the calculated velocities and angular accelerations into the satellite's EoMs instead of actual accelerations obtained from measurements, as shown in Figure \ref{fig:hilb_post}. 
As a result, current measurements were excluded from the setup, and only velocity measurements were used to interface with the simulator. 
Because the control and adaptation laws under test only require estimated satellite states and RWs' angular velocity measurements as inputs (as shown in Equations \eqref{eq:ctrl_law}, and \eqref{eq:adapt_law}), and EoMs propagation is only required for numerical simulation and not for real satellite operation, the proposed data exchange remains representative. 

However, the disconnection between the measured (experienced) RWs' angular velocities, and the calculated angular accelerations being fed into the EoMs, prevented us from physically inducing RW failures or degradation. In order to physically induce failure on a RW, an angular acceleration profile consistent with the measurement would be required. 
The need of angular acceleration is only present in this specific HIL test that requires EoMs propagation. However, in a more comprehensive test (e.g., tests involving a spherical air-bearing testbed), no artifacts would be required to construct RW angular acceleration and the framework proposed here remains fully applicable for physically induced failures.  Finally, to address the increased noise around zero RW speed, RWs were spun-up to an initial  $\boldsymbol{\Omega} = [100,-100, -100, 100]$ rad/s. With these adjustments, we achieved a closed-loop operation between the SatSim and the physical RWs.


\subsection{HIL-c: Adding Embedded Computer in the Loop}




In this stage, we aim to ensure that the controller is on the embedded hardware. 
This involved two additional tasks: (a) To check that the controller still provides correct output even when it receives the noisy velocity measurements, and (b) To ensure that the controller node correctly implements failure induction logic, i.e., computes the correct RW commands to be sent to the RW. 

Up to this point, RWs are still being commanded by the Satellite Simulator. As an intermediate step we executed the controller block on the Satellite Simulator and, in parallel, on the Jetson Nano embedded computer to assess their performance. 
Although the ControllerNode on the Jetson Nano had already been validated during the SIL stage, the main difference now is the noisy data measurements from the physical RWs, and newly incoporated satellite state estimates from an attitude determination EKF. 

The estimated states from the EKF (computed in Satellite Simulator) and real-time RW angular velocity measurements were sent to the ControllerNode on the Jetson Nano, and its outputs are transmitted back to the SatSimNode for comparison and verification. 
Once validated, we replace SatSim's RW commands with the commands calculated on the Jetson Nano to fully execute the control computations on the embedded computer. 


To validate the controller in this final setup, we employed the same test scenario from our previous work. 
In this scenario, the satellite begins by aligning itself with the Earth-Centered Inertial (ECI) frame and then alternates between its initial orientation and nadir-pointing three times. 
The simulation scenario lasts for 4000 seconds, during which the satellite switches its orientation every 12 minutes, then maintains nadir-pointing after 2000 seconds. 
The only changes we made to the simulation parameters were to increase the degraded wheel factor from 0 to 0.5 to induce a partial failure instead and reduce $K_{ICL}$ gain by an order of magnitude, i.e., ~$K_{ICL} = 1$. 
The gain adjustment was necessary to avoid demanding torque commands that exceed the limits of the RWs, which initially caused the simulator and the ControllerNode on the Jetson Nano to stop once the ICL term was activated. 
This highlights the gap in sim-to-real, where sometimes control torques that are feasible in simulation do not mean that they can be replicated in a real test. Table \ref{tab:gain_params} shows the gains that were adjusted to reflect the hardware constraints. 

\begin{table}[t!]
\centering
\caption{Controller Gains (Simulation vs. HIL)}
\begin{tabular}{l|c|c}
\toprule
\textbf{Gain} & \textbf{Simulation} & \textbf{HIL} \\
\midrule
$K_{\mathrm{ICL}}$ & $10$ & $1$ \\
$K$ & $5 \times 10^{-1}$ & $1 \times 10^{-2}$ \\
$\alpha$ & $3 \times 10^{-2}$ & $3 \times 10^{-2}$ \\
$\beta$ & $5 \times 10^{-3}$ & $5 \times 10^{-3}$ \\
$\gamma$ & $100I_{4}$ & $100I_{4}$ \\
$\bar{\lambda}$ & $1 \times 10^{-7}$ & $1 \times 10^{-7}$ \\
\bottomrule
\end{tabular}
\label{tab:gain_params}
\end{table}

\FloatBarrier
\section{Results}

\begin{figure}[h!]
    \centering
    \begin{subfigure}[t]{0.45\textwidth}
        \centering
        \includegraphics[width=\textwidth]{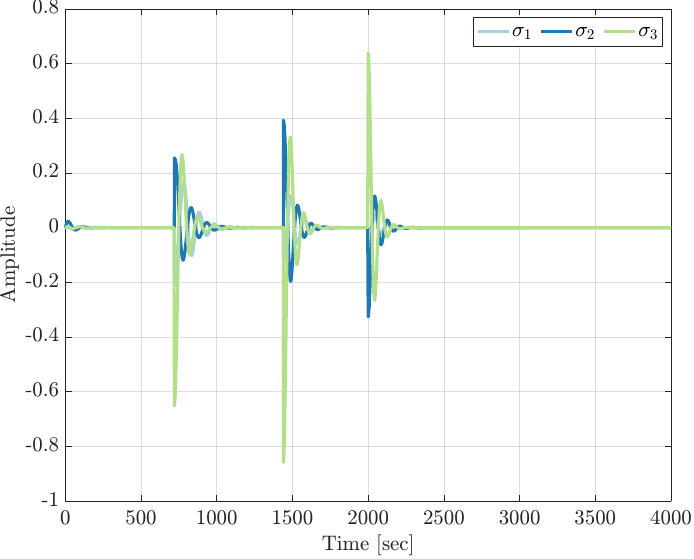}
        \caption{Error MRP}
        \label{fig:MRP}
    \end{subfigure}
    \hfill
    \begin{subfigure}[t]{0.45\textwidth}
        \centering
        \includegraphics[width=\textwidth]{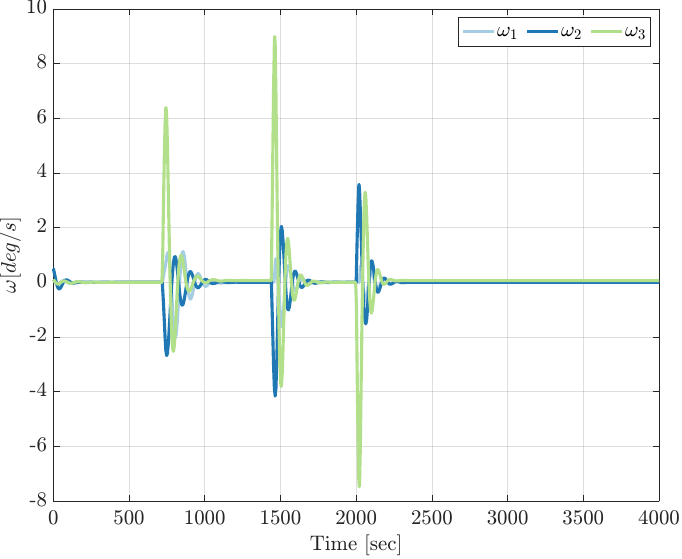}
        \caption{Body Angular Velocity}
        \label{fig:BAV}
    \end{subfigure}
    \caption{Performance of the Attitude Tracking Accuracy.}
    \label{fig:att_track}
\end{figure}

\begin{figure}[h!]
    \centering
    \begin{subfigure}[t]{0.45\textwidth}
        \centering
        \includegraphics[width=\textwidth]{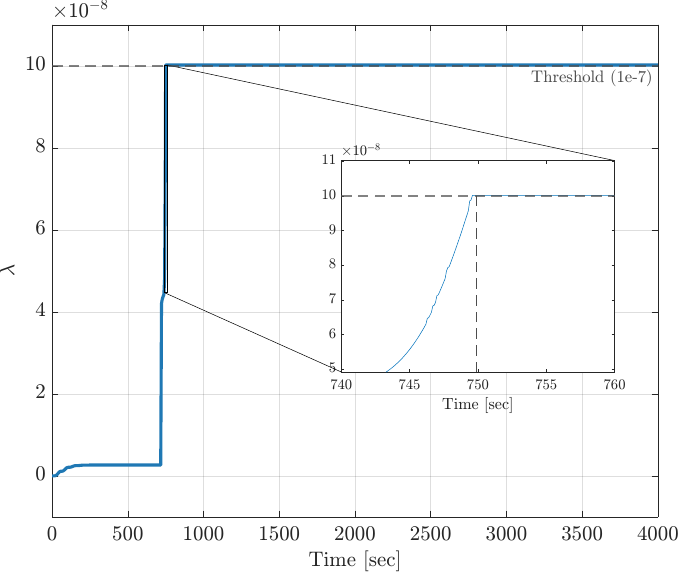}
        \small (a) Lambda 
        \label{fig:L}
    \end{subfigure}
    \hfill
    \begin{subfigure}[t]{0.45\textwidth}
        \centering
        \includegraphics[width=\textwidth]{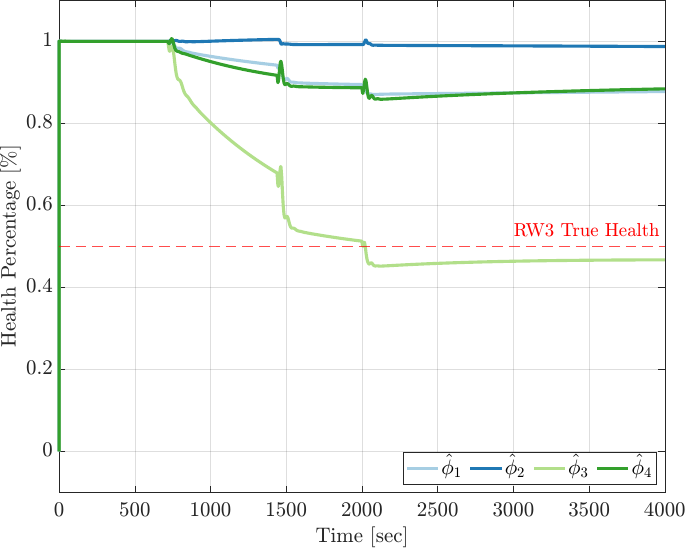}
        \small (b) Estimated RW Health
        \label{fig:PhiHat}
    \end{subfigure}
    \caption{RW Health Estimation with RW hardware in the loop.}
\label{fig:HIL_RW_Health}
\end{figure}

The satellite was able to follow its guidance commands throughout the simulation as evident in Figure \ref{fig:MRP} and Figure \ref{fig:BAV}. 
Figure \ref{fig:HIL_RW_Health} shows the RW health estimation performance, i.e., $\boldsymbol{\hat{\theta}}$. The lambda $\lambda$ plot represents the verifiable excitation level due to the input-output data accumulated by the system.\cite{nehma2025adaptive}
Although the lambda $\lambda$ reached the defined threshold value earlier than in purely simulated tests, the health estimation $\boldsymbol{\hat{\theta}}$ can be seen to converge to its true value. 
Because we reduce the gain ~$K_{ICL}$ to avoid the simulation from stopping, it affects the convergence rates, taking more time to converge. As the satellite perform more maneuvers, the action of the gradient-based term (i,e., first term in Equation \eqref{eq:adapt_law}) helped move it closer to its true value faster. 

Two consistent behaviors with prior results were observed. The estimation accuracy for degraded RW\#3 is better than for the non-degraded wheels (steady-state error) and that there was overshoot on the estimation for RW\#3. The steady state error was partly due to the numerical integration errors from the forward-Euler integration algorithm used to compute the estimated value $\boldsymbol{\hat{\theta}}$ via integration of Equation \eqref{eq:adapt_law}, as well as the fact that in the development of the controller some parts of the simulation model, such as attitude perturbations were neglected, leading to a disparity between the two. 

\begin{figure}[h!]
    \centering
    \includegraphics[width=0.5\linewidth]{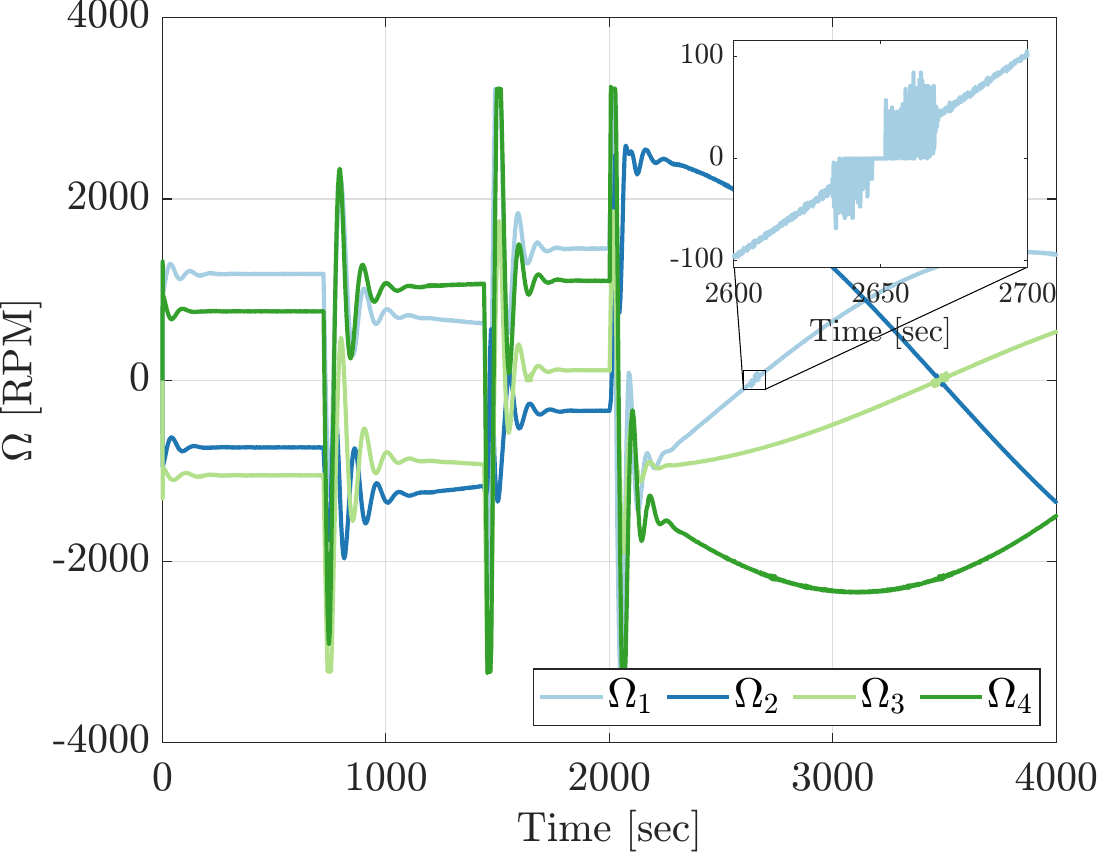}
    \caption{RW Angular Velocity Measurement.}
    \label{fig:HIL_RW_Ang_Vel}
\end{figure}

Figure \ref{fig:HIL_RW_Ang_Vel} shows the angular velocity measurements from the RW. 
Due to internal low-pass filtering implemented in the digital motor controller, the measurement noise was observed only at low-speed regions. 
Despite the measurement noise, it has minimal impact on the performance of the attitude tracking and estimation of the controller, as seen in earlier plots. 

Table \ref{tab:jetson_nano_perf} reports the performance of the Jetson Nano embedded computer during the HIL test. 
The controller executes the loop in $3.86 \pm 0.20$ \si{ms} on average, which is well below the 100 \si{ms} of the control loop limits. 
To measure the ROS2 nodes CPU load consumption, each node was assigned to a specific CPU. The ControllerNode, running on CPU2, consumed less than $15$\%, while the RWNode, running on CPU3, consumed about $10$\% of CPU load. 
The maximum CPU load of CPU2 and CPU3 reached $100$\%, which is expected when the nodes are initialized at the start of the HIL test. 
The remaining CPU0 and CPU1 handle the operating system and background tasks that, on average, consume less than 3\% of CPU load. 
For memory usage, it increased to 19.6\% with the controller running if compared to the baseline of 16.6\%, which means that only a small amount of memory (3\%) is required to execute the control algorithms. 

\begin{table}[t!]
\centering
\caption{Jetson Nano Performance Metrics.}
\begin{tabular}{lccc}
\toprule
\textbf{Metric} & \textbf{Average} & \textbf{Maximum} & \textbf{Std. Dev.} \\
\midrule
Execution Time (ms) & 3.86 & 6.62 & 0.20 \\
\addlinespace[2pt]
\multicolumn{4}{l}{\textit{CPU Usage (\%)}} \\
\quad CPU2 (ControllerNode) & 14.7 & 100 & 4.4 \\
\quad CPU3 (RWNode)         & 10.3 & 100 & 5.0 \\
\quad CPU0                  & 2.7  & 6.90 & 0.9 \\
\quad CPU1                  & 0.3  & 3.00 & 0.5 \\
\addlinespace[2pt]
\multicolumn{4}{l}{\textit{Memory (RAM \%)}} \\
\quad Idle (baseline) & 16.6 & 16.7 & 0.01 \\
\quad HIL             & 19.6 & 19.8 & 0.6 \\
\bottomrule
\end{tabular}
\label{tab:jetson_nano_perf}
\end{table}

\FloatBarrier
\section{Future Work}
Given the results and lessons learned in this HIL testing, the avenues for future work are highly promising. The next immediate step in validating the adaptive controller in a simulation setup that mimics its true mission profile as closely as possible is to transition the RW hardware setup from a flat test bench array to one where the RWs are placed in the design configuration array inside a mock satellite on a rotational air bearing. This setup would alleviate the problematic issues caused by the EoM in the simulation, allowing for a more comprehensive and thorough analysis of the controller. 

With this setup, another possibility for future work includes moving the EKF that is currently hosted on the simulation computer, along with the attitude determination node, onto the embedded computer, as it would be on a real mission. This testing would ensure all necessary communication and computations that would be required by the embedded computer would be possible without causing too much computational burden. 

Finally, with the EoM now being omitted from the simulation setup, we would then be able to physically induce failures to the RW again to an even more realistic scenario to test the learning capability of the controller. With this component of the HIL test, performed on the air bearing testbed, full validation of the adaptive controller would be achieved.

\section{Conclusion}
In this paper, we validated an adaptive satellite attitude controller capable of estimating the health level of its RWs through the HIL test that includes an embedded computer and physical RW hardware with artificially induced failure.
Using the HIL testbed, we demonstrated that the proposed adaptive controller performed considerably well under hardware constraints, with noisy measurements from the sensor.
In addition, through the test, we have established a baseline validation of our envisioned modular HIL testbed design that is flexible and scalable. 
Despite several issues encountered during the development of the HIL testbed, the lessons learned are valuable as we move towards the next step involving a full air-bearing testbed experiment and other comprehensive end-to-end system tests in the future. 
This work will contribute to advancing the state-of-the-art ground hardware testing of fault-tolerant spacecraft attitude controllers.

\section{Acknowledgment}
The authors would like to thank Cole Schumacher for his contribution to the initial tests and integrations of the RWs to the HIL testbed. 

\bibliographystyle{AAS_publication}   
\bibliography{referencesHIL}   

\end{document}